# Anomalous Frame Detection Using VLM-Based Description Comparison for Extracting Expert-Specific Actions and Contextual Decision-Making Scenes with Intra-Video Self-Similarity

**Ryo Sakai[1] and Kaname Yokoyama[1]**

[1] Robotics Research Department, Research and Development Group, Hitachi, Ltd., Ibaraki, Japan

Corresponding author: Ryo Sakai (e-mail: ryo.sakai.cx@hitachi.com).

**ABSTRACT** Maintenance of critical infrastructures, such as railways and power plants, is essential for ensuring operational safety and reliability. However, the declining number of skilled maintenance workers highlights the need to transfer expert know-how to less experienced workers. Previous studies have attempted to extract candidates of expert knowledge by comparing videos of manual-based work with those of expert workers, mainly focusing on differences in observable actions. However, expert know-how is often embedded not only in actions but also in contextual decision-making during task execution. This paper proposes a method that detects anomalous frames between two task videos to automatically extract candidate scenes containing expert-specific actions and contextual decision-making scenes. The method generates frame-wise visual descriptions using a vision–language model (VLM). Expert-specific actions are extracted based on frame similarities computed from description comparisons between two videos, while contextual decision-making scenes are extracted using segment similarities derived from intra-video self-similarity of the descriptions. In simulated distribution board maintenance experiments involving 27 task scenarios, the proposed method achieved extraction rates of 65% for action candidates and 61% for decision-scene candidates, improving over a previous method that achieved 59% and 33%, respectively. These results demonstrate the effectiveness of the proposed approach in discovering candidate scenes containing expert know-how.

**INDEX TERMS** Anomalous frame detection, video similarity analysis, vision–language models, expert actions, expert contextual decision-making, maintenance skill transfer

## I. INTRODUCTION

Maintenance work is one of the critical tasks in the operation of infrastructure facilities such as railways and power plants. On the other hand, there is a shortage of skilled maintenance personnel with sufficient technical skills at each site [1, 2, 3]. One potential solution to this shortage is the training of new maintenance inspectors. For new inspectors to acquire know-how equivalent to that of experts, it takes time to re-experience the work performed by those experts. For example, cases have been reported where expert know-how in ship operations is used for the early training of personnel [4]. However, as expert know-how is often tacit, its codification is essential for systematic education and efficiently expanding the pool of inspectors [5]. Therefore, there is a growing need to extract work know-how from experts, codify it as explicit knowledge, and pass it on to new maintenance personnel.

One method for codifying the work know-how of skilled maintenance workers (hereinafter referred to as "experts") is the interview technique [4, 6, 7, 8, 9, 10, 11]. However, it is difficult to obtain information regarding know-how that the interviewer cannot anticipate in advance or that the expert themselves is unaware of. Furthermore, work know-how that is second nature to the expert or that they do not

* This work was carried out as part of an internship by Kaname Yokoyama; he is currently with Graduate school of Engineering, Toyota Technological Institute.

perceive as important may not be uncovered. Since such know-how may actually underpin the quality of the work, it is desirable to codify it, including this unconscious know-how.

To codify know-how that even experts are unaware of, the conventional method extracts differences as action candidates that may contain know-how by generating and comparing image descriptions for each video—one created based on a manual (hereinafter referred to as a "manual-based work video") and the other recording the expert's work (hereinafter referred to as an "expert's work video") [12]. This technology has demonstrated that actions added by experts or skipped relative to the manual can be detected as differences in behavior corresponding to "what to do" in work procedures.

In addition to such "what to do" actions, expert know-how can also include contextual decision-making, such as "where and which item to operate," even when performing the same task. For example, consider a case where a manual-based work video records a breaker-off task for a certain distribution board during maintenance, and an expert's work video records the same task for a different distribution board. In this case, the task is the same breaker-off operation, so no difference in action occurs. However, the decision regarding which distribution board to operate may reflect the expert's advanced know-how concerning system configuration, load status, safety, and other factors.

In this way, even if the action is the same, differences in the work scenes such as the target or location—may contain decision-related know-how. Therefore, it is important to extract decision-making scenes that may contain such know-how. However, since the conventional method [12] generates descriptions independently for each image in each video, the focus of description generation may not align between videos, and different descriptions may be generated for individual frames even for the same task. Consequently, differences in descriptions may not necessarily reflect differences in decision-making scenes, but may arise from differences in visual focus, making it difficult to extract appropriate decision-making scenes.

A simple way to realize the extraction of decision-making scenes while suppressing the influence of attention shifts in the conventional method [12] is to input all image pairs from each video into an image description generator, directly compare the images, and output descriptions of whether the scenes are different to extract candidates. With this method, the image description generator can determine if a scene is a decision-making one while referring to both images and aligning the focus between them. However, in this approach, descriptions for image pairs that are temporally close are expected to be similar. Yet the generator may produce different descriptions depending on minute differences between images transitioning over time, potentially leading to lower extraction accuracy based on incorrect descriptions. Furthermore, direct comparison between images increases the computational complexity on the order of the square of the number of images, thereby increasing calculation time.

Therefore, this study proposes a method that uniformly realizes the extraction of action candidates and decision-making scene candidates within a single framework. Specifically, action candidates are obtained through a filtering process using a cross-video frame similarity map created from image descriptions generated from each frame. For decision-making scene extraction, each video is first segmented into semantically stable intervals using a self-similarity map, and different scenes are then extracted by comparing image pairs within these segments. A key feature of the proposed method is that it realizes the extraction of both action and decision-making scene candidates in a unified manner and expands the comparison unit for decision-making scene extraction from an image-wise to a segment-wise basis. This approach suppresses the reduction in extraction accuracy caused by description instability and mitigates the impact of increased computational complexity.

In experiments involving 27 task scenarios of simulated distribution board maintenance, the proposed method achieved extraction rates of 65.0% for action candidates and 61.0% for decision-scene candidates. These results mark a significant improvement over the conventional method, which achieved 59.0% and 33.0% for action and decision-scene candidates, respectively. These findings demonstrate that the proposed method can effectively support the codification of expert know-how by improving extraction accuracy while maintaining computational efficiency comparable to existing approaches.

The main contributions of this paper are summarized as follows:

i. A method for automatically extracting candidate expert-specific actions and contextual decision-making scenes from task videos. We formulate the problem of identifying candidate expert actions and contextual decision-making scenes by detecting anomalous frames between manual-based work videos and expert worker videos, all without the need for additional training.
ii. A unified framework that integrates frame-level and segment-level video comparisons based on VLM-generated descriptions. Expert-specific actions are extracted using frame similarities computed from VLM-based description comparisons between two videos, while contextual decision-making scenes are extracted using segment similarities derived from intra-video self-similarity analysis.
iii. Experimental validation on simulated distribution board maintenance tasks. Experiments on 27 simulated maintenance scenarios demonstrate that the proposed method achieves extraction rates of 65% for action candidates and 61% for decision-scene candidates, outperforming conventional methods (59% and 33%).

The remainder of this paper is organized as follows. Section II reviews related work on video difference detection. Section III details problem definition. Section IV details the proposed method, including similarity map generation and video segmentation. Section V and VI explain the experimental design and results. Finally, Section VII presents the conclusion of this study.

## II. RELATED WORK

Various methods have been proposed to compare two videos and extract their differences. For example, some approaches compute similarities between two videos to construct a similarity map and detect anomalous frames based on this map [13, 14, 15, 16]. These methods generally consist of two main steps: (1) constructing a similarity map and (2) detecting differences using the similarity map. By employing deep learning–based techniques, they can robustly detect crucial differences even in the presence of noise that should not be the focus, such as variations in camera angles and lighting conditions.

In addition, studies have been proposed that utilize large language models (LLMs) and vision–language models (VLMs) to extract task-related differences between two videos [17]. In these approaches, the level of abstraction of the differences to be detected can be specified as a text query, enabling flexible adjustment of the detection process. These methods directly detect differences by taking two videos as input.

However, many of these methods require large amounts of video data and training based on the collected data. Although the cost of data collection can be reduced in domains where data are readily available on the web, in domains such as maintenance and inspection, where data are generally not publicly available, the cost of data collection and training becomes high. Therefore, for practical applications, it is important to detect differences with as low cost as possible.

To address this issue, this paper proposes a difference detection method that utilizes similarity maps constructed from descriptions generated by pre-trained VLMs and captioning models, aiming to minimize the cost of deployment in real environments.

Related to this study, there has been growing research on video understanding and video comparison using VLMs. Recent studies have explored the use of pre-trained VLMs for video understanding tasks [18]. For example, ActionCLIP [19] reformulates action recognition as a matching problem between videos and text and achieves zero-shot transfer through prompt-based fine-tuning. VideoLLM [20] employs a decoder-only LLM to perform unified inference across multiple video tasks. These studies demonstrate the potential of leveraging pre-trained models to understand video content with minimal additional training.

Furthermore, VideoDiff [21] has been proposed as a method for detecting differences between two videos without additional training. VideoDiff uses multiple pre-trained models to generate natural language descriptions of differences between two input videos, optionally guided by textual descriptions of the video content. However, while VideoDiff handles temporal information regarding when differences occur, it does not explicitly identify difference segments. Moreover, it focuses on differences in actions, such as variations in form or skill within the same action and does not address differences in work scenes arising from contextual decision-making.

In our previous work [12], we proposed a difference detection method based on similarity maps constructed from VLM-based captioning texts between two videos. This method was shown to be effective in extracting anomalous frames related to expert action knowledge. However, since the method is based on frame-level difference detection, it is difficult to capture differences in entire work scenes arising from contextual decision-making. This study extends this framework to handle both action-level and scene-level differences in a unified manner.

Another related research area is scene change detection, which aims to detect differences in scenes between two videos. Scene change detection compares a reference video and a target video to identify visual changes such as object appearance/disappearance and background structural changes. Padilla et al. [22] proposed a method that synchronizes two videos temporally and computes feature differences to detect newly appearing objects. Recently, training-free methods have also been proposed. Cho et al. [23] detect changes by combining object detection results from a pre-trained model with tracking-based identity management. Kim et al. [24] proposed a change detection method based on pseudo-mask generation using internal features of pre-trained models and two-stage matching using geometric and semantic information.

These methods [22, 23, 24] mainly focus on detecting local visual changes such as object appearance and disappearance. In contrast, the anomalous decision-scene detection targeted in this study aims to extract segments in which the overall work scene differs between a manual-based video and an expert video, due to differences in expert decision-making, such as changes in operation targets, inspection targets, movement destinations, or gaze points. The differences considered in this study may include local changes such as object appearance/disappearance but are not limited to them. Rather, the key objective is to detect task-semantic scene differences that reflect differences in decision-making within the work context, rather than pixel-level or object-level differences. Furthermore, while existing scenes change detection methods typically output change masks or regions, our method outputs video segments containing anomalous candidates. In addition, these methods do not address action anomalies and decision-scene anomalies within a unified framework.

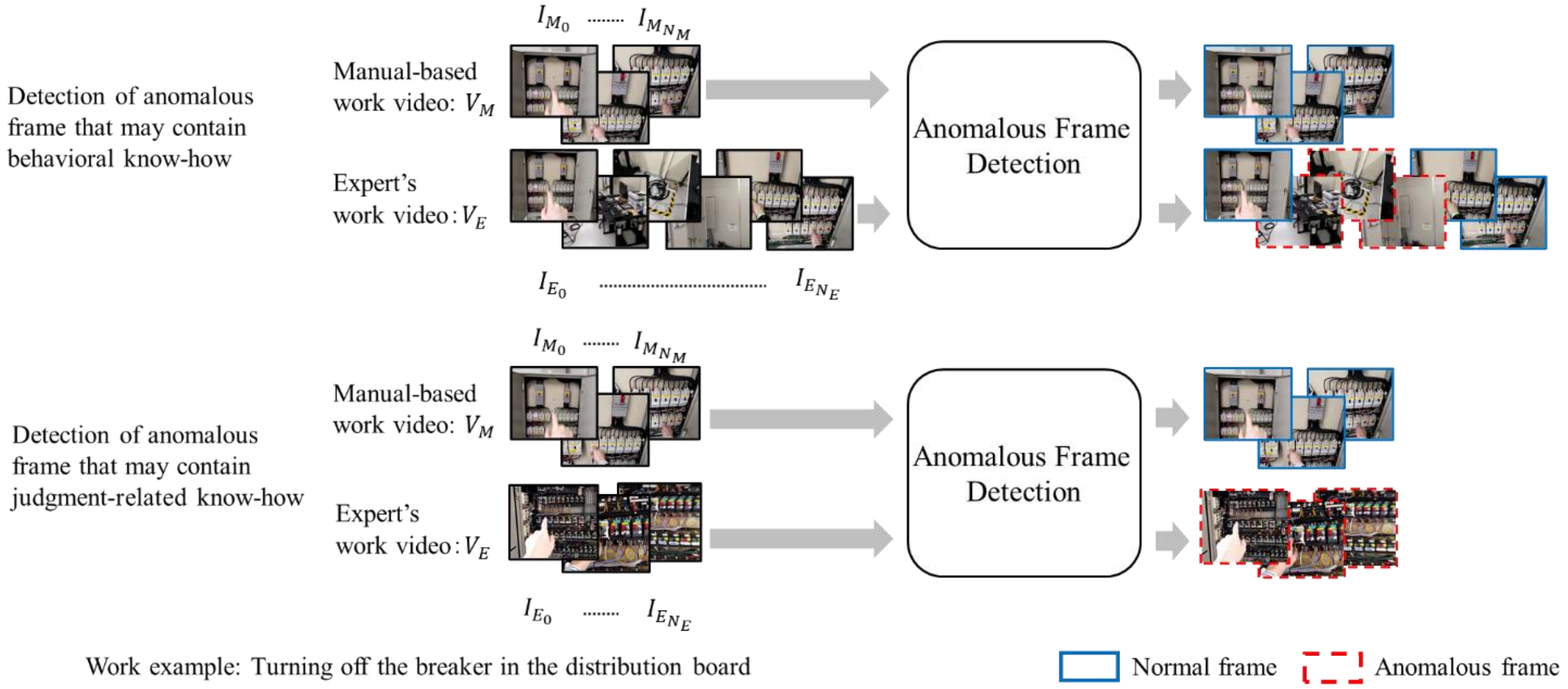

**FIGURE 1. Conceptual flow diagram illustrating the process of extracting candidate actions and candidate scenes containing expert know-how from both the manual-based work video $V_M$ and the expert's work video $V_E$.**

Therefore, this paper proposes a method that directly compares a manual video and an expert video to extract both candidate action anomalies and candidate decision-scene anomalies in a unified and training-free manner, contributing to the extraction of expert know-how.

## III. PROBLEM DEFINITION

The objective of this study is to automatically extract candidate actions or candidate decision scenes that may contain expert know-how from a manual video and an expert video. Here, a *decision scene* refers to a work scene that may contain know-how related to the expert's judgment, for example, a scene in which, even for the same power distribution panel shutdown task, the expert selects which panel to shut down based on the on-site situation and their own experience. Figure 1 shows a conceptual diagram of the flow for extracting candidate actions or candidate decision scenes that may contain expert know-how from the above two videos.

Let the manual video be denoted by $V_M = \{I_{M_i} \mid 1 \le i \le N_M\}$, and the expert video by $V_E = \left\{I_{E_j} \middle| 1 \le j \le N_E\right\}$. Here, $I_{M_i}$ is the image of the $i$-th frame constituting $V_M$ , and $I_j^E$ is the image of the $j$-th frame constituting $V_E$. Both $V_M$ and $V_E$ are first-person-view videos captured from the worker's perspective.

In this study, to extract candidate actions or candidate decision scenes that contain expert know-how, we propose a method that takes $V_M$ and $V_E$ as input and detects the start frame $F_s$ and end frame $F_e$ of the interval in which such candidates appear. After detecting the start frame $F_s$ and end frame $F_e$, candidate actions or candidate decision scenes that may contain expert know-how can be extracted by retrieving the images between these frames (the frames enclosed by the red dashed line in Figure 1; hereafter referred to as *anomalous frames*). The remaining images (the frames enclosed by the blue solid line in Figure 1; hereafter referred to as *normal frames*) are considered to correspond to work performed by the expert in accordance with the manual, and are therefore treated as not containing expert know-how.

Figure 2 shows examples of $V_M$, $V_E$, and the target start frame $F_s$ and end frame $F_e$ to be detected. Figure 2(a) assumes, as a task example, shutting off the power of a distribution panel breaker after work, and illustrates the case where $V_E$ contains (A) an action additionally performed because the expert deemed it necessary. $V_M$ is a work video created based on the manual for the task, in which the worker performs the breaker shutdown described in the manual. The additional action in $V_E$ is the action of checking the outlet box connected to the distribution panel before shutting off the breaker power. These frames belong to the anomalous frames in $V_E$ (the red dashed frames in Figure 2(a)). In such a case where an additional action exists, the goal is to detect the start frame $F_s$ and end frame $F_e$ for the anomalous frame interval in $V_E$.

Figure 2(b) assumes, as a task example, a patrol inspection of the work room conducted before shutting off the power of a distribution panel breaker and illustrates the case where (B) an action present in the manual is not performed in $V_E$. Compared with $V_M$, the expert in $V_E$ skips the action of confirming that the door is closed.

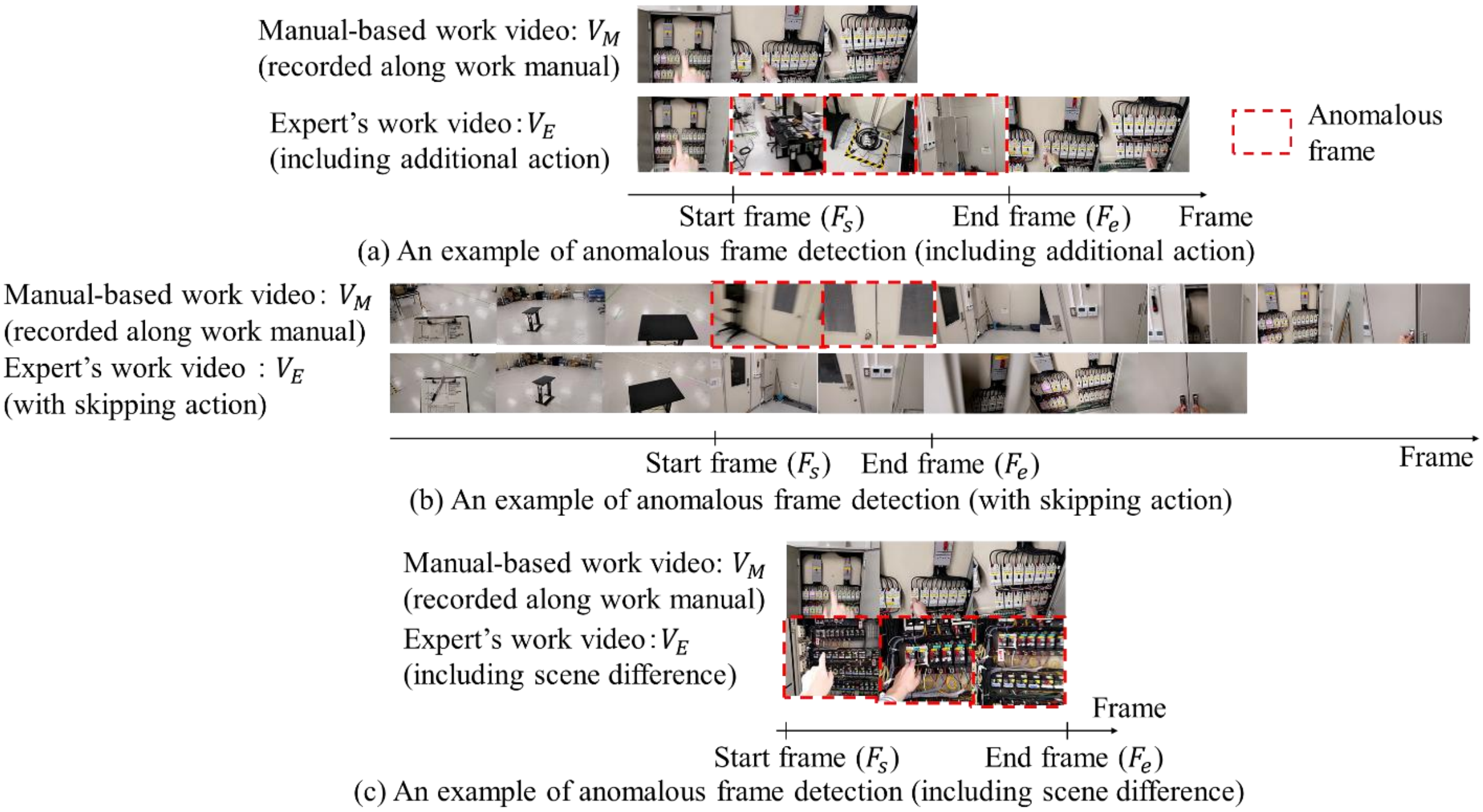

**FIGURE 2. Examples of the manual-based work video ($V_M$), the expert's work video ($V_E$) and the start frame ($F_s$) and end frames ($F_e$) to be detected. (a) An example of anomalous frame detection (including additional action). (b) An example of anomalous frame detection (with skipping action). (c) An example of anomalous frame detection (including scene difference).**

This action belongs to the anomalous frames in $V_M$ (the red dashed frames in Figure 2(b)). In such a case where an action is skipped, the goal is to detect the start frame $F_s$ and end frame $F_e$ within the anomalous frames of $V_M$.

Figure 2(c) assumes the same breaker shutdown task as in Figure 2(a) and illustrates the case where $V_E$ contains (C) a work scene involving a different distribution panel from that in $V_M$, based on the expert's situational judgment. From the viewpoint of task content, $V_M$ and $V_E$ involve the same task, and thus no action difference arises. On the other hand, from the viewpoint of the work scene, $V_M$ and $V_E$ differ, and therefore the corresponding frames belong to the anomalous frames in $V_E$ (the red dashed frames in Figure 2(c)). In such a case where there exists an interval in which the work scene differs, the goal is to detect the start frame $F_s$ and end frame $F_e$ for the anomalous frame interval in $V_E$.

In Figure 2(c), the entire video corresponds to a different work scene; however, there are also cases where only part of the video corresponds to a different scene. For example, the outlet box confirmation in Figure 2(a) can be interpreted as an action resulting from the expert's situational judgment regarding the work scene. Even in such a case where only part of the video corresponds to a different work scene, the goal is similar to detect the start frame $F_s$and end frame $F_e$ for the anomalous frame interval in $V_E$.

## IV. PROPOSED METHOD

### *A. Overview*

Figure 3 shows an overview of the proposed method for detecting the start frame $F_s$ and end frame $F_e$ of intervals containing candidate actions and decision scenes that may include expert know-how, from a manual video $V_M$ and an expert video $V_E$. The proposed method detects the start frame $F_s$ and end frame $F_e$ by comparing $V_M$ and $V_E$ and determining whether there exists a portion depicting an action that appears only in one of the videos (hereafter referred to as an *anomalous action*). It also detects the start frame $F_s$ and end frame $F_e$ by comparing $V_M$ and $V_E$ and determining whether there exists a portion depicting a scene that appears only in one of the videos (hereafter referred to as an *anomalous scene*). To realize this detection scheme, the proposed method performs anomalous-action detection in two steps and anomalous-scene detection also in two steps.

In the first step of anomalous-action detection, a similarity map is generated whose elements are the similarities between pairs of images constituting $V_M$ and $V_E$, respectively (the green block in Figure 3). To compute the similarity between images, the method uses a vision-language model (VLM) and a captioning instruction prompt to generate a description for each image and then computes the similarity between the generated descriptions. The captioning instruction prompt specifies that the purpose of the VLM is to caption the action-related content in the video.

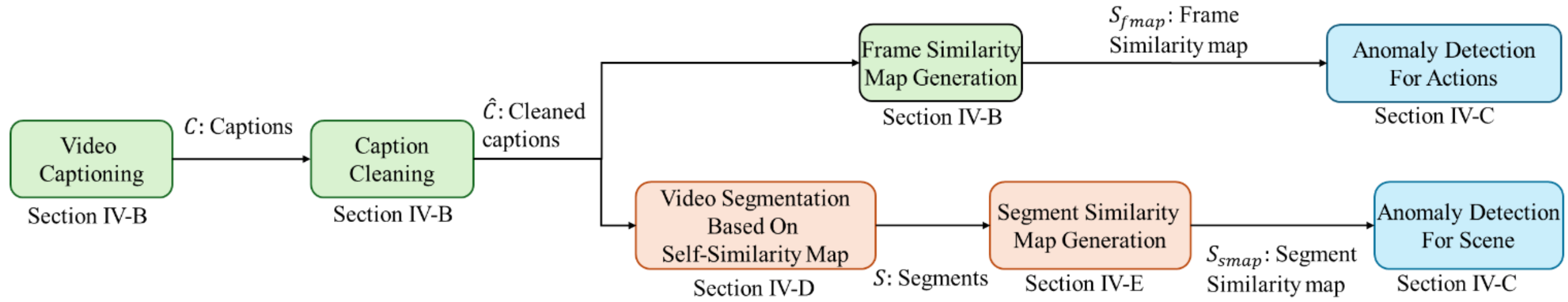

**FIGURE 3.** Conceptual flowchart of the proposed method.

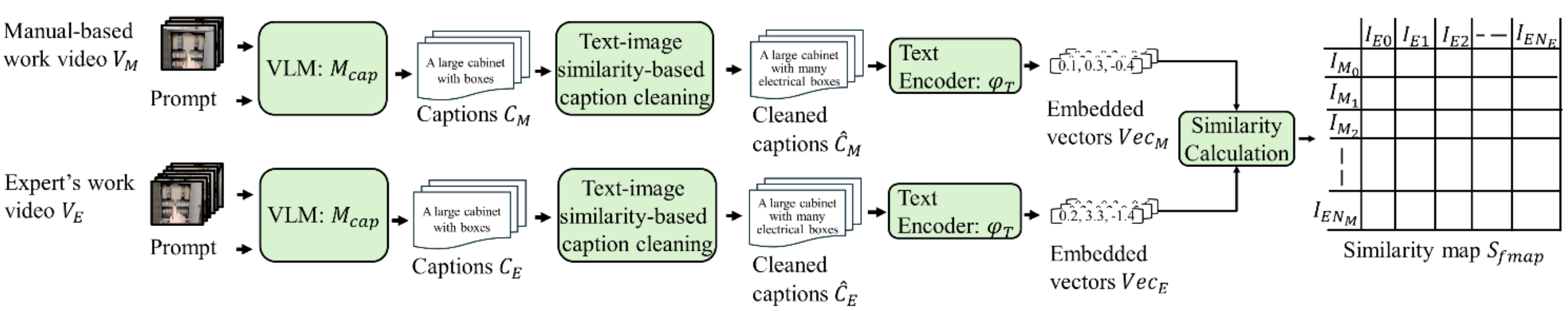

**FIGURE 4.** Workflow for creating similarity maps between videos $V_M$ and $V_E$. For each frame of the videos, the proposed method generates image captions $C_M$ and $C_E$ using a VLM $M_{\text{cap}}$. To improve the quality of the generated descriptions, the method vectorizes images and captions, compares their similarities to select the closest caption, and uses it as the cleaned caption. Then, the method vectorizes the cleaned captions $\widetilde{C}_M$ and $\widetilde{C}_E$ using a text encoder $\phi_T$, calculates the similarity between each pair of vectors $\text{Vec}_M$ and $\text{Vec}_E$, and creates a frame similarity map $S_{\text{fmap}}$ whose elements are the calculated similarities.

In this similarity map, when similar actions are performed in the two videos, regions with high similarity are expected to appear. On the other hand, when an anomalous action exists only in one of the videos, regions with low similarity to all images in the other video are expected to appear.

Therefore, in the second step of anomalous-action detection, the method identifies regions with low similarity in the similarity map and detects portions depicting actions that exist only in one of the videos (the blue block in Figure 3). Specifically, filtering is applied to the similarity map to determine whether the similarity satisfies a certain condition, and the beginning and end of the frames satisfying the condition are detected as the start frame $F_s$ and end frame $F_e$, respectively.

In the first step of anomalous-scene detection, each of $V_M$ and $V_E$ is first segmented into video segments, and a similarity map is then created whose elements are the similarities between pairs of segments (the orange block in Figure 3). In video segmentation, the similarities between the descriptions of images within each video, which were generated in the first step of anomalous-action detection, are computed to construct a self-similarity map for each individual video. In this self-similarity map, regions with high similarity are expected to appear for similar frames whose descriptions are alike, whereas regions with low similarity are expected to appear for dissimilar frames. Since neighboring frames in a work video are generally expected to be similar, segmenting the video according to diagonal blocks in the self-similarity map makes it possible to divide the video into groups of frames with semantically similar descriptions. Because these segments can be interpreted as portions of the video separated at a certain semantic granularity, comparing segments between $V_M$ and $V_E$ enables comparison between the videos at an appropriate level of granularity.

Accordingly, in the second step of anomalous-scene detection, the similarities between the segments of $V_M$ and $V_E$ created in the first step are computed to construct a segment-level similarity map, and processing similar to that in the second step of anomalous-action detection is performed (the blue block in Figure 3). In this study, the beginning and end of the frames satisfying a specific condition for being regarded as an anomalous scene are detected as the start frame $F_s$ and end frame $F_e$, respectively.

Section IV-B explains how to generate the similarity map between images in $V_M$ and $V_E$. Next, Section IV-C explains the filtering method for detecting anomalous frames from the similarity map. Section IV-D describes the video segmentation method based on the self-similarity map constructed from the image descriptions in Section IV-B. Finally, Section IV-E explains how to create a similarity map between video segments.

### *B. Generation of a Similarity Map Between Videos*

Figure 4 illustrates the procedure for constructing a similarity map between $V_M$ and $V_E$. First, captions are generated for the images, and the captions are vectorized using a text encoder. Then, a similarity map is constructed by computing similarities between the vectors corresponding to the images. Although it is also possible to vectorize images directly using an image encoder, such an approach is expected to be strongly affected by changes in images that are not directly related to the task, such as differences in camera angle or lighting conditions. Therefore, to achieve robustness to such variations and to vectorize task-relevant content in the images, the proposed method performs captioning for each image and generates vectors representing the captions.

Since the processing for extracting candidate actions is described in detail in the previous method [12], only the outline and the main improvements are explained here. First, captions corresponding to the frames constituting a video are generated using a caption-capable VLM ($M_{\mathrm{cap}}$) according to Eq. (1):

$$C = \{c_i = M_{cap}(I_i, P_{cap}) | \ 0 \leq i \leq N\} \tag{1}$$

Here, $N$ is the number of frames constituting the video, and $c_i$ is the caption corresponding to the $i$-th frame. $P_{\mathrm{cap}}$ is a captioning instruction prompt, and is an improvement newly introduced in this study over the previous method [12] to improve the accuracy of similarity-map generation by controlling the content of the captions. $C$ denotes the set of captions $c_i$, i.e., the captions for all images constituting the video. This caption generation is performed for the frames of both $V_M$ and $V_E$, producing the caption sets $C_M$ for $V_M$ and $C_E$ for $V_E$. The same captioning model is used for both $V_M$ and $V_E$.

Next, low-quality captions for a given image are corrected by replacing them with a more appropriate caption selected from among the captions assigned to other images, thereby generating a corrected caption set $\tilde{C}$. The outline of the construction of $\tilde{C}$ is given by Eq. (2):

$$\tilde{c_i} = \arg \max_{c \in C} \langle \varepsilon_I(I_i), \varepsilon_T(c) \rangle \tag{2}$$

Here, $\varepsilon_I$ and $\varepsilon_T$ are the image encoder and text encoder, respectively, jointly trained in an image-text foundation model, and the latent vectors $\varepsilon_I(I_i)$ and $\varepsilon_T(c)$ output by these encoders can be treated as vectors in the same space. Therefore, by computing the latent vector $\varepsilon_I(I_i)$ of image $I_i$ and comparing it with the latent vectors $\varepsilon_T(c)$ of all captions $c \in C$, the caption most similar to the image is regarded as the corrected caption $\tilde{c}_i$, thereby correcting low-quality captions. In Eq. (2), $\langle \cdot,\cdot \rangle$ denotes the similarity measure, for which cosine similarity is used. By constructing $\tilde{c}_i$ for each image, the corrected caption sets $\tilde{C}_M$ and $\tilde{C}_E$ are obtained.

Next, the corrected caption sets $\tilde{C}_M$ and $\tilde{C}_E$ are converted into sets of latent vectors, denoted by $\mathrm{Vec}_M$ and $\mathrm{Vec}_E$, using the text encoder $\phi_T$. The cosine similarity is then computed between every pair of latent vectors, and a frame-level similarity map $S_{fmap}$ is constructed whose elements are these similarities. $S_{fmap}$ is an $N_M \times N_E$ matrix, and the element in the $i$-th row and $j$-th column of $S_{fmap}$ represents the similarity between the $i$-th frame image $I_i^M$ of $V_M$ and the $j$-th frame image $I_j^E$ of $V_E$.

### C. Filtering of Anomalous Frames Based on the Similarity Map

In the similarity map $S_{fmap}$ created in Section IV-B, the similarities of images corresponding to (A) actions additionally performed in $V_E$ because the expert independently judged them necessary, and (B) actions present in the manual but not performed in $V_E$, are relatively smaller than those of other images. Exploiting this property, the proposed method computes the average maximum similarity in $S_{fmap}$, and detects the start frame $F_s$ and end frame $F_e$ by comparing the maximum similarity in each column or each row with this average maximum similarity.

Algorithm 1 shows the filtering procedure for anomalous-frame detection based on the similarity map. In Algorithm 1, the case where anomalous frames exist in $V_E$, i.e., the case where $V_E$ contains (A) an action additionally performed because the expert judged it necessary, is assumed for explanation.

First, the average maximum similarity $MaxAve$ of $S_{fmap}$ is computed by Eq. (3). Each column of $S_{fmap}$ corresponds to a frame in $V_E$, and each element of the similarity vector $S_{fmap,k}$ in the $k$-th column indicates how similar the $k$-th image $I_k^E$ in $V_E$ is to each image in $V_M$. If the expert is performing work in accordance with the manual, then an image corresponding to $I_k^E$ should exist among the images constituting $V_M$, and therefore one of the similarity values in $S_{fmap,k}$ should be high; ideally, if the same image exists, that similarity should become the maximum value in $S_{fmap,k}$ Based on this observation, the proposed method adopts the mean of the maximum values of the column-wise similarity vectors in $S_{fmap}$ as a criterion for judging whether $I_k^E$ is an anomalous frame, and defines it as the average maximum similarity $MaxAve$:

$$MaxAve = \frac{\sum_{l=0}^{N_E} \max(S_{map,k})}{N_E} \tag{3}$$

Next, Eq. (4) is used to determine whether $I_k^E$ is an anomalous frame. The threshold $\tau$ is a coefficient multiplied by $MaxAve$, and controls how sensitively the proposed method reacts to differences between the two videos, where $\tau > 0$. The larger $\tau$ is, the more likely the method is to regard even a slight decrease in similarity as an anomalous frame and extract it as a candidate action containing expert know-how. If the maximum similarity of $S_{fmap,k}$ is less than or equal to the value obtained by multiplying the average maximum similarity $MaxAve$ by the threshold, then $I_k^E$ is regarded as an anomalous frame and recorded in the anomalous frame set $F_{anomalous}$:

$$\max(S_{map,k}) \leq \tau \cdot MaxAve \tag{4}$$

After this judgment is performed for all columns of $S_{fmap}$, the first frame in $F_{anomalous}$, obtained by the function GetFirstValue($F_{anomalous}$), is regarded as the start frame $F_s$. Similarly, the last frame in $F_{anomalous}$, obtained by the function GetEndValue($F_{anomalous}$), is regarded as the end frame $F_e$.

When anomalous frames exist in $V_M$, i.e., in case (B) where an action present in the manual is not performed, the start frame $F_s$ and end frame $F_e$ can likewise be detected by changing the scanning direction of Algorithm 1 from column-wise to row-wise. In this way, by applying the processing of Algorithm 1 in both the row and column directions of $S_{fmap}$, the proposed method detects frames depicting candidate actions that may contain expert know-how, as described in Section III.

The direction of the above filtering process is determined as follows: when $N_M < N_E$, the difference is regarded as case (A), i.e., an additional action by the expert, and the filtering is performed in the column direction of $S_{fmap}$; otherwise, it is regarded as case (B), i.e., an action skipped by the expert, and the filtering is performed in the row direction.

Figure 5 shows a conceptual diagram of filtering applied to $S_{\mathrm{fmap}}$ using Algorithm 1. Figure 5(a) shows the case where $V_E$ contains (A) an action additionally performed because the expert independently judged it necessary, and Figure 5(b) shows the case where anomalous frames exist in $V_M$, i.e., (B) an action present in the manual is not performed. For the $S_{\mathrm{fmap}}$ shown in Figure 5, the average maximum similarity is computed by Eq. (3), and the anomalousity of each frame is sequentially judged by Eq. (4), as indicated by the arrows in Figures 5(a) and 5(b).

**Algorithm 1** Average filtering process for detecting anomalous frame

**Require:** $S_{fmap}$, $\tau$
**Ensure:** $F_s$, $F_e$

$$MaxAve \Leftarrow \frac{\sum_{l=0}^{N_E} \max\left(S_{fmap,l}\right)}{N_E} \quad (3)$$

$F_{anomalous} = \phi$
**for** $k = 0$ to $N_E$ **do**
  **if** $\max\left(S_{fmap,k}\right) \leq \tau * MaxAve$ (4) **then**
    $F_{anomalous} = F_{anomalous} \cup k$
  **end if**
**end for**
$F_s \Leftarrow GetFirstValue(F_{anomalous})$
$F_e \Leftarrow GetEndValue(F_{anomalous})$

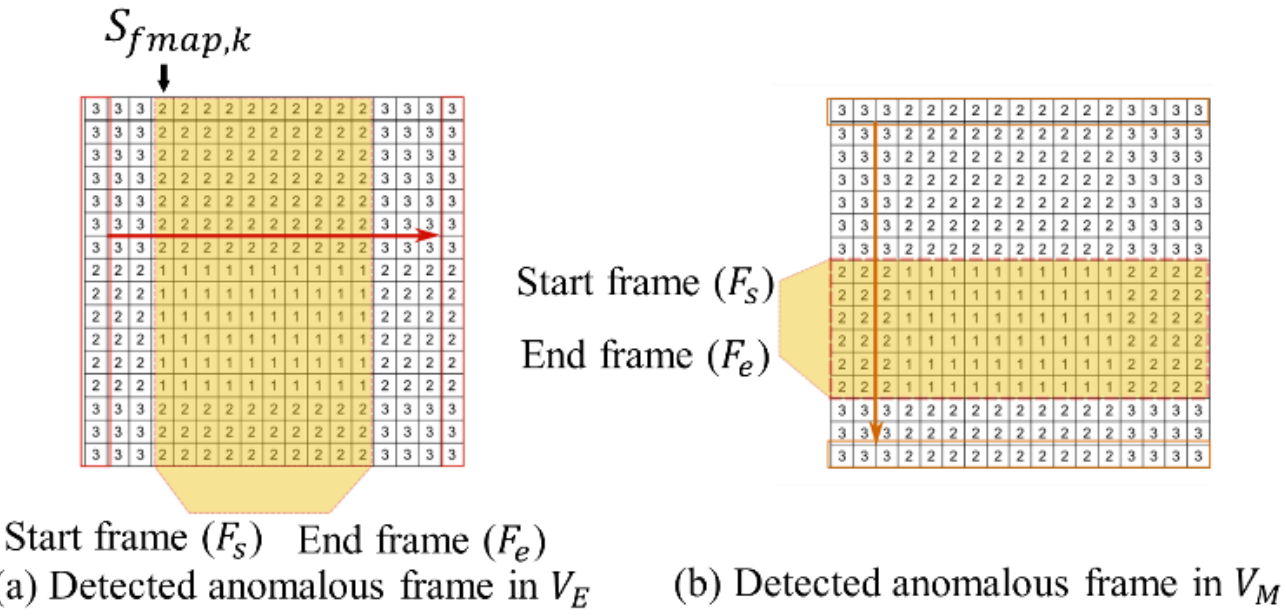

**FIGURE 5. Conceptual diagram of filtering applied to $S_{\mathrm{fmap}}$. (a) An example of detected anomalous frames in $V_E$. (b) An example of detected anomalous frames in $V_M$.**

### *D. Video Segmentation Based on a Self-Similarity Map*

The similarity map $S_{fmap}$ created in Section IV-B can quantify the frame-to-frame correspondence between the manual video $V_M$ and the expert video $V_E$, and is therefore effective for extracting candidate actions contained only in one of the videos. On the other hand, even when the action itself is the same, differences in the work scene, such as the work target or work location, may contain know-how related to the expert's judgment. However, in the previous method [12], image descriptions are generated independently for each video, and thus the focus of the description generation is not necessarily aligned across videos. As a result, differences between descriptions may be caused not by differences in judgment-relevant work scenes but by mismatches in attention.

To avoid this problem, one possible approach is to input pairs of images, one from $V_M$ and one from $V_E$, into a VLM and directly compare image pairs. However, this approach has the drawback that the number of comparisons increases and the computational cost grows quadratically. In addition, a description generator that takes image pairs as input may generate different descriptions depending on subtle differences between temporally adjacent images, which may lead to degraded extraction accuracy due to erroneous decision-scene extraction based on incorrect descriptions.

Therefore, in this study, the videos are segmented into semantically stable intervals based on the self-similarity map of each video, and inter-video comparison is performed at the segment level, thereby achieving both robustness and computational efficiency in decision-scene extraction. In addition, candidate action extraction is realized, as in the previous method, by filtering a frame-level similarity map, and is handled within a unified framework together with decision-scene extraction.

First, as shown in Figure 6(a), cosine similarities are computed between the latent vectors in $\mathrm{Vec}_M$, which correspond to the corrected caption set $\tilde{C}_M$ defined in Eq. (2) in Section IV-B, to create a similarity map $S_{self,M}$ whose elements are similarities between image descriptions within a single video. $S_{self,M}$ is an $N_M \times N_M$ matrix, and the element in the $i$-th row and $j$-th column represents the similarity between the $i$-th frame image $I_i^M$ and the $j$-th frame image $I_j^M$ in $V_M$. The same procedure is applied to $V_E$ to create $S_{self,E}$. $S_{self,E}$ is an $N_E \times N_E$ matrix, and the element in the $i$-th row and $j$-th column represents the similarity between the $i$-th frame image $I_i^E$ and the $j$-th frame image $I_j^E$ in $V_E$.

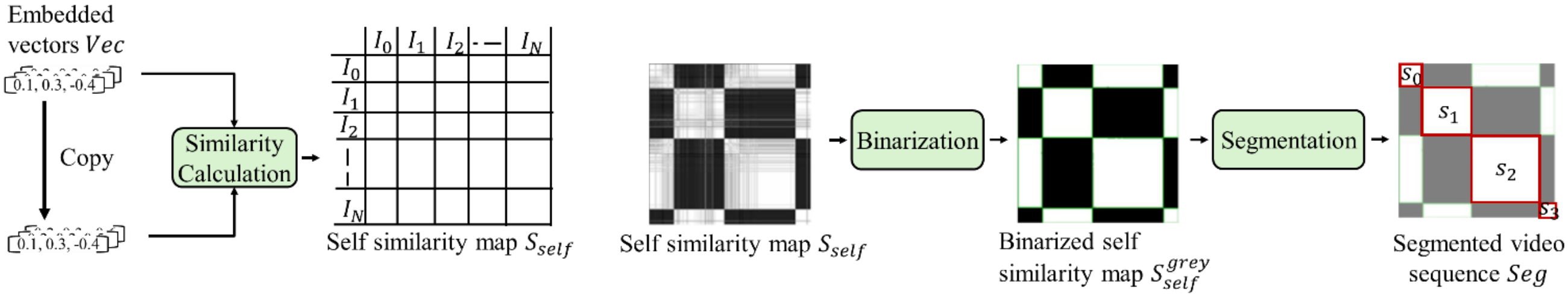

**FIGURE 6. Video segmentation process based on a self-similarity map. (a) Calculation of self-similarity map $S_{self}$. (b) Binarization of the self-similarity map and video segmentation.**

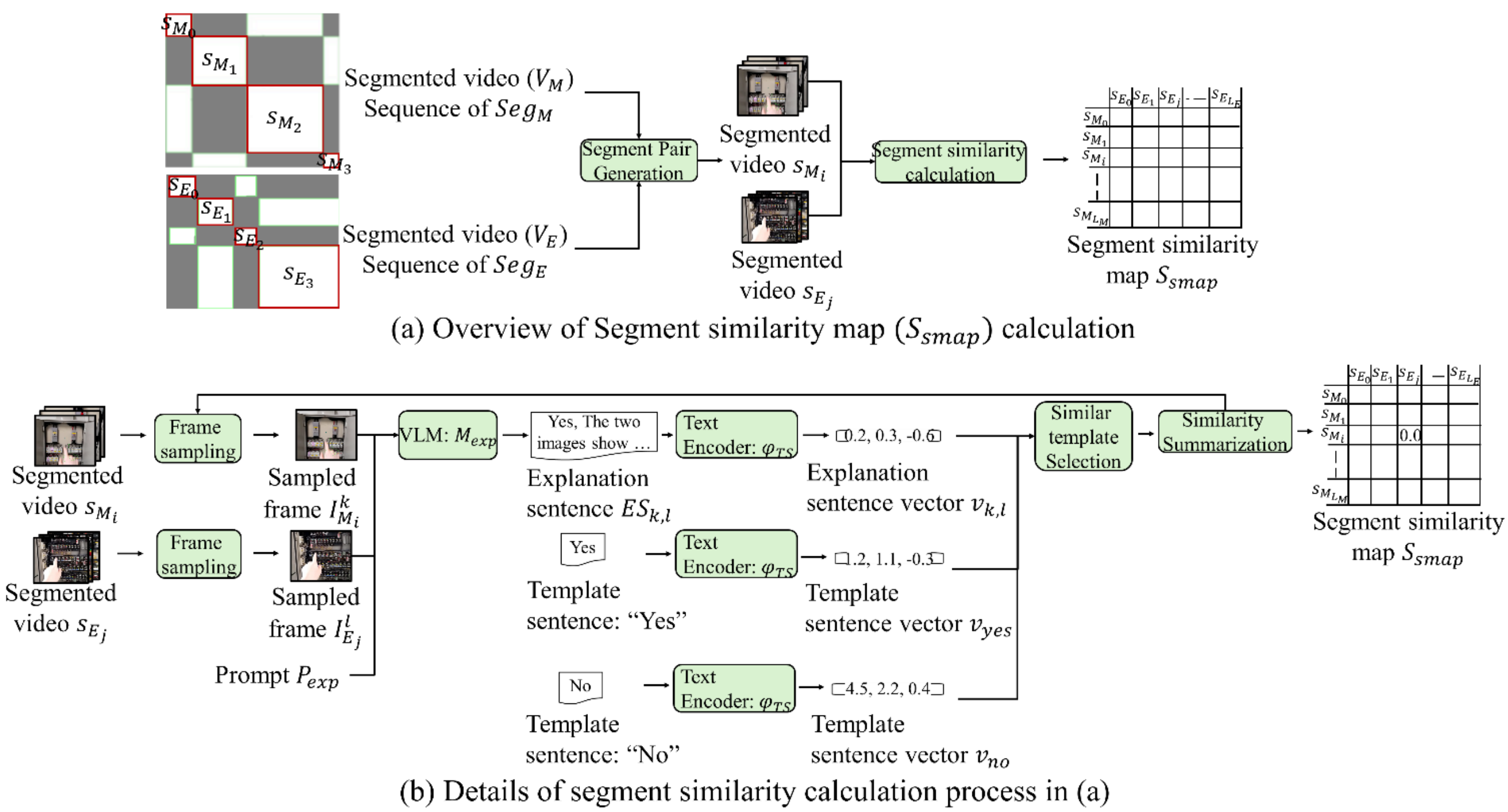

**FIGURE 7. Generation process of the segment similarity map. (a) Overview of segment similarity map $S_{smap}$. (b) Details of the segment similarity calculation process in (a).**

Next, as shown in Figure 6(b), $V_M$ and $V_E$ are segmented according to the diagonal blocks of the self-similarity maps $S_{self,M}$ and $S_{self,E}$. To clearly distinguish diagonal blocks composed of highly similar elements from the other regions in the self-similarity maps, the maps are treated as grayscale images and binarized using a threshold, producing binary maps $S_{self,M}^{grey}$ and $S_{self,E}^{grey}$, in which highly similar diagonal blocks appear white and the other regions appear black. Then, the Connected Component method [25] is applied to $S_{self,M}^{grey}$ and $S_{self,E}^{grey}$, respectively, to extract connected components of white regions. Among the extracted connected components, those located on the diagonal and having an area greater than or equal to a given threshold are regarded as a sequence of video segments, denoted by $Seg$. $Seg$ is defined by Eq. (5):

$$Seg = [s_i = [F_i^s, F_i^e] \mid 0 \le i \le L] \tag{5}$$

Here, for each segment $s_i$, the start frame $F_i^s$ and end frame $F_i^e$, corresponding to the extent of the component along the diagonal direction, are obtained and defined as the boundaries of the video segment. By computing $Seg_M$ and $Seg_E$ for $V_M$ and $V_E$, respectively, multiple semantically stable video segments are obtained for each video. Section IV-E describes how similarities are computed for all combinations of these segments to create a segment-level similarity map.

### E. Construction of a Segment-Level Similarity Map

Using $Seg_M$ and $Seg_E$, which are the segment sequences created for $V_M$ and $V_E$ in Section IV-D, a similarity map $S_{smap}$ is constructed whose elements are the similarities between pairs of video segments. Figure 7(a) shows an overview of the generation process of $S_{smap}$.

As shown in Figure 7(a), all possible pairs are formed between the segments constituting $Seg_M$ and those constituting $Seg_E$, and the similarity for each pair is calculated.

Figure 7(b) shows the details of the similarity calculation for a pair of video segments. First, frames are sampled from segment $s_{M_i}$, which is a segment constituting $Seg_M$, to obtain images $I^k_{M_i}$. Similarly, frames are sampled from segment $s_{E_j}$, which is a segment constituting $Seg_E$, to obtain images $I^l_{E_j}$. Next, using a VLM ($M_{exp}$) that can accept multiple images as input, a scene-consistency description $ES_{k,l}$, indicating whether $I^k_{M_i}$ and $I^l_{E_j}$ depict the same scene, is generated according to Eq. (6):

$$ES_{k,l} = M_{exp}(I^k_{M_i}, I^l_{E_j}, P_{exp}) \tag{6}$$

Here, $P_{exp}$ is a prompt instructing the model to determine whether the two input images belong to the same scene. Specifically, $P_{exp}$ instructs the model to answer "Yes" if the two images depict the same scene, and "No" otherwise. However, depending on the prompt-following ability of the VLM ($M_{exp}$), the model may produce variations of these answers, such as alternative expressions conveying "Yes" or "No", or additional explanations following "Yes" or "No". Therefore, it is necessary to introduce a mechanism for automatically determining whether the free-form scene-consistency description $ES_{k,l}$ indicates that the two images belong to the same scene.

To this end, the proposed method compares the latent vector $v_{k,l}$, corresponding to the scene-consistency description $ES_{k,l}$, with template latent vectors $v_{\text{yes}}$ and $v_{\text{no}}$, corresponding to template sentences explicitly indicating "Yes" and "No," using a text encoder $\phi_{TS}$. Specifically, the dissimilarity probability $P(k,l)$ of $v_{k,l}$ with respect to the template sentences is computed according to Eq. (7):

$$P(k,l) = \frac{\exp(\langle v_{k,l}, v_{\text{no}} \rangle)}{\exp(\langle v_{k,l}, v_{\text{no}} \rangle) + \exp(\langle v_{k,l}, v_{\text{yes}} \rangle)} \tag{7}$$

As in Eq. (2), cosine similarity is used for comparing latent vectors, and the similarity operation is denoted by $\langle \cdot,\cdot \rangle$. If the dissimilarity probability $P(k,l)$ is higher than the threshold $th_{\text{exp}}$, the two images are judged not to belong to the same scene, and their similarity is set to 0. Otherwise, the two images are judged to belong to the same scene, and their similarity is set to 1.

This frame sampling process and judgment process are performed at equal intervals for an arbitrary number of frames within $s_{M_i}$ and $s_{E_j}$. The results are then integrated such that, if even one of the sampled comparisons yields a similarity of 0, the similarity of the segment pair is set to 0; otherwise, it is set to 1. This integrated value is used as the similarity between the two segments, i.e., as each element of the segment-level similarity map $S_{smap}$. Finally, the filtering process described in Section IV-C is applied to the constructed $S_{smap}$. In decision-scene extraction, the filtering is performed in the column direction of $S_{smap}$ in order to detect the start frame $F_s$ and end frame $F_e$ within the anomalous frames of $V_E$.

## V. EXPERIMENT

### A. Overview

In this study, using maintenance tasks for distribution panels as the target scenario, we created a work video based on a manual and videos of expert work, and conducted experiments to automatically extract candidate actions and candidate decision scenes that may contain expert know-how. First, a worker performed the task according to the manual, and a first-person-view video $V_M$ was recorded during the work. Next, the worker performed the task while adding actions corresponding to expert behavior and recorded a video $V_E$ that included (A) actions additionally performed because the expert independently judged them necessary. In addition, for experiments assuming the formalization of expert behavior in which redundant actions described in the manual are appropriately skipped, the worker performed the task without executing (B) certain actions in the manual and recorded the resulting video $V_E$. Furthermore, for experiments aimed at detecting differences in decision scenes as a basis for formalizing the expert's situational judgment, the worker recorded a video $V_E$ containing work performed in a different scene. In this experiment, $V_M$ and $V_E$ were compared to extract candidate actions and candidate decision scenes that may contain expert know-how.

### B. Collected data

In this experiment, data were collected for the following three maintenance tasks related to distribution panels: (task1) breaker shutdown work for a distribution panel, (task2) inspection work of a work room before a distribution panel shutdown operation, and (task3) disassembly work of a management PC for a distribution panel.

For task1, as shown in $V_M$ in Figure 2(a), the following actions were defined as manual-based actions and recorded in $V_M$: checking the distribution panel breaker and turning the breaker OFF. In contrast to these actions, as shown in $V_E$ in Figure 2(a), the following were added as (A) expert actions and recorded in $V_E$: checking the connected devices in the outlet box connected to the breaker before turning the breaker OFF (A-1), and checking related equipment below the distribution panel before turning the breaker OFF (A-2). The recording of $V_M$ and $V_E$ for (A-1) in task1 was conducted in two different buildings (Building 1 and Building 2). In Building 1, one worker collected one $V_M$ and three $V_E$ videos, and another worker collected one $V_E$ video. In Building 2, two workers each collected one $V_M$ and one $V_E$. In addition, the recording of $V_M$ and $V_E$ for (A-2) in task1 was conducted in Building 1, where one worker collected one $V_M$ and one $V_E$. As a result, a total of seven pairs of $V_M$ and $V_E$ were collected for the extraction experiment of candidate expert actions in task1.

In addition, as expert videos $V_E$ containing decision scenes different from those in $V_M$, as shown in Figure 2(c), we collected (c-1) expert videos in which the work scene differs throughout the entire video, and (c-2) expert videos in which only part of the work scene differs from $V_M$. The videos in (c-2) further include two cases: one in which the manual-based action and the expert action are the same, but part of the work scene differs, and another in which the expert performs an additional action, resulting in a partial change in the work scene. For (c-1), we assumed the same breaker shutdown task but targeting a different distribution panel. One pair was collected in which the work video from Building 1 was treated as $V_M$ and the work video from Building 2 was treated as $V_E$, and two pairs were collected in which the work video from Building 2 was treated as $V_M$ and the work video from Building 1 was treated as $V_E$. For (c-2), we assumed tasks in which part of the work scene in $V_E$ differs from $V_M$ because one of the following actions is added: checking the connected devices in the outlet box connected to the breaker before turning the breaker OFF (c-2-1), opening the door of the distribution panel (c-2-2), or turning on a lighting switch (c-2-3). Three pairs of $V_M$ and $V_E$ containing action (c-2-1) in $V_E$ were collected; however, one of these pairs was also used as a pair for the extraction experiment of candidate expert actions, so only two pairs were newly collected. In addition, one pair containing action (c-2-2) in $V_E$ and one pair containing action (c-2-3) in $V_E$ were collected. As a result, a total of eight pairs of $V_M$ and $V_E$ were collected for the extraction experiment of candidate expert decision scenes in task1.

For task2, as shown in $V_M$ in Figure 2(b), the following were defined as manual-based actions and recorded in $V_M$: checking the desk, door, air-conditioner power supply, and distribution panel in that order. In contrast to these actions, the following were added as (A) expert actions and recorded in $V_E$: actually, moving to the air conditioner and confirming that it was not operating (A-1), and storing equipment if equipment was present on the desk (A-2). The recording of $V_M$ and $V_E$ for task2 was conducted in Building 1, and one $V_M$, one $V_E$ including action (A-1), and one $V_E$ including action (A-2) were collected for each of two workers. In addition, as a video recording expert behavior of type (B), a $V_E$ video was recorded in which the action of checking the door was not performed. For this type-(B) recording, one worker collected one $V_E$, and the other worker collected two $V_E$ videos. As a result, a total of seven pairs of $V_M$ and $V_E$ were collected for the extraction experiment of candidate expert actions in task2.

In addition, as expert videos $V_E$ containing decision scenes different from those in $V_M$, we collected, as in task1, (c-1) expert videos in which the work scene differs throughout the video, and (c-2) expert videos in which only part of the work scene differs. For (c-1), we first assumed the same inspection task but in a different target room. One pair was collected in which the work video from Building 1 was treated as $V_M$ and the work video from Building 2 was treated as $V_E$. We also assumed the same inspection task under a different lighting environment and collected one pair in which a work video recorded under a normal environment was treated as $V_M$, and a work video recorded in a dark environment without lighting was treated as $V_E$.

For (c-2), we assumed tasks in which part of the work scene in $V_E$ differs from $V_M$ because one of the following is added: a task in which equipment is stored when equipment is present on the desk (c-2-1), or a task in which the worker continues the work while ignoring the equipment left on the desk (c-2-2). One pair of $V_M$ and $V_E$ containing action (c-2-1) in $V_E$ was collected; however, this pair was also used as a pair for the extraction experiment of candidate expert actions and was therefore not newly collected. One pair of $V_M$ and $V_E$ containing action (c-2-2) in $V_E$ was also collected. As a result, a total of four pairs of $V_M$ and $V_E$ were collected for the extraction experiment of candidate expert decision scenes in task2.

For task3, as manual-based actions, the sequence of separating a desktop PC used for distribution panel management into the CPU cooler, GPU, and memory, in that order, was defined and recorded in $V_M$. In contrast, as a video recording expert behavior of type (B), a $V_E$ video was recorded in which the memory was not removed. One worker collected one $V_M$ and one $V_E$. As a result, a total of one pair of $V_M$ and $V_E$ was collected for the extraction experiment of candidate expert actions in task3.

Consequently, a total of 15 pairs were collected across tasks 1 to 3 for the extraction experiment of candidate expert actions. In addition, a total of 12 pairs were collected across tasks 1 to 3 for the extraction experiment of candidate expert decision scenes. The operators wore a Microsoft Hololens2 integrated helmet [26] for data collection. The operators recorded data via a modified StreamRecorder application from Hololens2ForCV library [27], modified to start and stop recording and to display holograms.

### C. Metrics

For each collected $V_M$ and $V_E$, the start frame $F_{S_{gt}}$ and end frame $F_{E_{gt}}$ of each action and each work scene in the video were manually annotated to create ground-truth data. In this study, to evaluate the automatic extraction of candidate actions and candidate decision scenes that may contain expert know-how, we used the metric shown in Eq. (8), which calculates how well the interval defined by the extracted start frame $F_S$ and end frame $F_E$ covers the ground-truth interval defined by the start frame $F_{S_{gt}}$ and end frame $F_{E_{gt}}$. Eq. (8) was computed for each collected pair of data, and the average value over all pairs was used for evaluation.

$$Acc_{union} = \frac{(F_E - F_S + 1) \cap \left(F_{E_{gt}} - F_{S_{gt}} + 1\right)}{(F_E - F_S + 1) \cup \left(F_{E_{gt}} - F_{S_{gt}} + 1\right)} \quad (8)$$

In the extraction experiment for candidate actions, for pairs in which the type of candidate action was incorrectly estimated, such as estimating an action addition when the action was actually skipped or estimating an action skip when the action was actually added, the value of Eq. (8) was set to 0. In addition, the extraction time for each pair was also measured.

To confirm the effectiveness of the proposed method, comparative experiments with conventional methods were conducted. In addition to the previous method [12], a method in which prompt instructions were given to the VLM used in the previous method [12] (hereafter referred to as previous method + instruction prompt) was evaluated in order to examine whether prompt instructions to the VLM increase or decrease extraction performance. Furthermore, in order to confirm the effectiveness of the video segmentation and segment-level similarity-map-based decision process of the proposed method, a method that directly compares each image of the videos by simultaneously inputting them into an image-description generator (hereafter referred to as the naive method) was also evaluated. In addition, as a simple method for comparing two videos, a method that compares two videos using a video-capable VLM (hereafter referred to as the video comparison method) was also evaluated.

To select the VLM ($M_{cap}$) used in the proposed method, the following models were tested in the candidate-action extraction experiment: BLIP-2 [28], InstructBLIP [29], InternVL2 [30], LLaVA [31], and Qwen3-VL-Instruct [32], which had also been used in the previous method. InternVL2, which achieved the best performance, was adopted (details are described in Section VI). In addition, the text encoder $\phi_T$ used for candidate-action extraction in Section IV-B was ImageBind [33]. The threshold $\tau$ in Algorithm 1 was set to 0.6 with reference to the previous method [12]. The binarization process used for video segmentation in Section IV-D employed the binarization function provided by OpenCV [25], with the threshold set to 100. Furthermore, the text encoder $\phi_{TS}$ described in Section IV-E was SentenceBERT [34], and the threshold $th_{exp}$ was set to 0.5. The sampling process was performed five times. In addition, Qwen3-VL-Instruct, which can perform inference on multiple images, was used as $M_{exp}$.

The conventional methods and the proposed method differ in their approaches to detecting anomalous frames. In addition, the inference time required for anomalous frame detection depends on both the detection method and the combination of VLMs employed. Since shorter inference times are preferable in practical applications, this experiment measures the inference time of each method in addition to evaluating detection performance. The experiments were conducted on a workstation equipped with an NVIDIA RTX 6000 Ada GPU.

## VI. RESULTS

Figure 8 shows an example of the frame-level similarity map $S_{fmap}$ generated by the proposed method from $V_M$ and $V_E$ for task1. As shown in Figure 8, the action of turning the distribution panel breaker OFF is present in both $V_M$ and $V_E$, and the corresponding region in the similarity map $S_{fmap}$ is confirmed to have high similarity. In this way, actions that are present in both $V_M$ and $V_E$ exhibit high similarity in the corresponding regions of $S_{fmap}$. In contrast, the action of checking the connected devices in the outlet box connected to the breaker before turning the breaker OFF does not exist in any part of $V_M$, and therefore the corresponding region shows low similarity when the similarity map $S_{fmap}$ is viewed in the column direction. These results confirm that the similarity map $S_{fmap}$ generated by the proposed method can represent regions depicting candidate actions that may contain expert know-how in the manual-based work video and the expert work video.

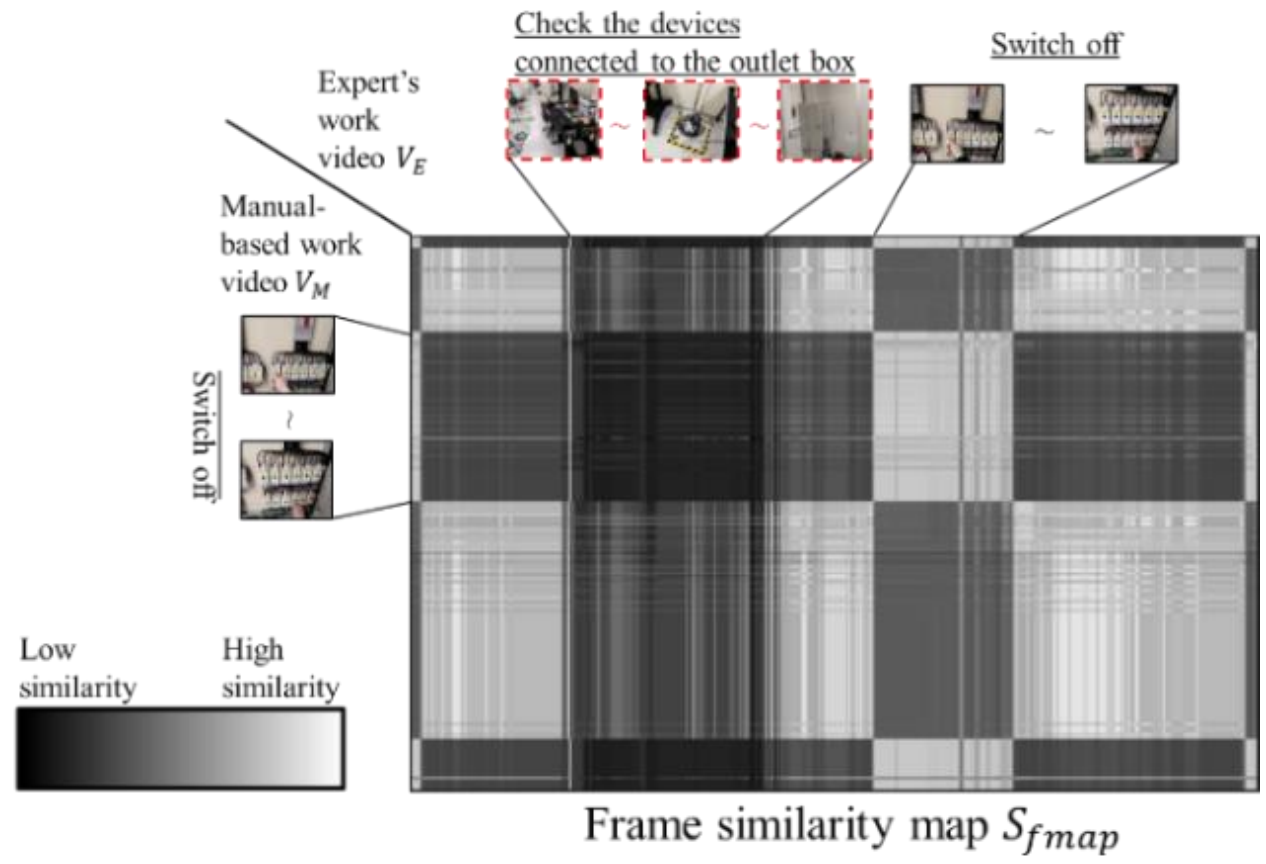

**FIGURE 8.** Example of the frame similarity map $S_{fmap}$ generated from $V_M$ and $V_E$ using the proposed method.

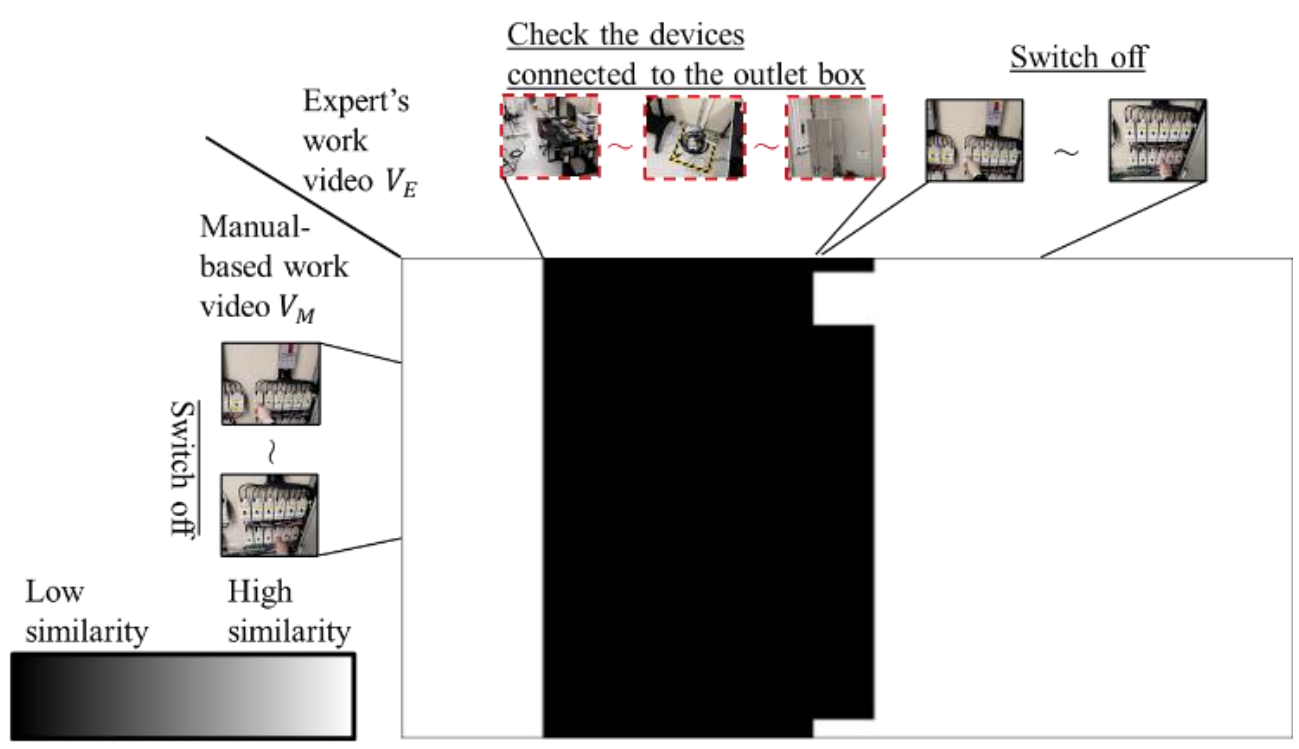

**FIGURE 9. Example of the segment similarity map $S_{\text{smap}}$ generated from $V_M$ and $V_E$ using the proposed method.**

TABLE I
EXTRACTION RESULTS OF CANDIDATE ACTIONS FOR THE VLM USED IN THE PROPOSED METHOD

| VLM | Params. of VLM | $Acc_{union}$ | Inference time |
|---|---|---|---|
| BLIP-2 | 2.7B | 53.0 % | 3 min |
| InstructBLIP | 7B | 44.0 % | 8 min |
| LLaVA | 7B | 63.0 % | 7 min |
| InternVL2 | 4B | 65.0 % | 6 min |
| Qwen3-VL-Instruct | 8B | 49.0 % | 5 min |

Figure 9 shows an example of the segment-level similarity map $S_{smap}$ generated by the proposed method from $V_M$ and $V_E$ for task1. As shown in Figure 9, for the same work scene in $V_M$ and $V_E$, the corresponding region in the similarity map $S_{smap}$ is confirmed to have high similarity. On the other hand, checking the connected devices in the outlet box connected to the breaker before turning the breaker OFF causes a change in the work scene, and since this work scene does not exist in any part of $V_M$, the corresponding region shows low similarity when the similarity map $S_{smap}$ is viewed in the column direction. These results confirm that the segment-level similarity map $S_{smap}$ generated by the proposed method can represent regions depicting candidate decision scenes that may contain expert know-how in the manual-based work video and the expert work video.

Table 1 shows the action-candidate extraction results obtained with different VLMs used in the proposed method. As shown in Table 1, the extraction accuracy of candidate actions varies depending on the VLM used. In this experiment, InternVL2, which achieved the highest accuracy, was adopted as the VLM ( $M_{\text{cap}}$ ) for candidate-action extraction.

Table 2 shows the evaluation results on the collected data using the proposed method and the baseline methods. As shown in the action-candidate extraction results in Table 2, the proposed method (d) achieved an accuracy of 65.0% for extracting candidate actions that may contain expert know-how, which is 6 percentage points higher than that of the previous method (a). It can also be seen that the proposed method extracts candidate actions containing expert know-how more accurately than the other baseline methods, namely the naive method (c) and the video comparison method (b). In terms of computation time, the proposed method (d) required several minutes, which is of the same order as that of the previous method (a).

In addition, as shown in the decision-scene candidate extraction results in Table 2, the proposed method (m) achieved an accuracy of 61.0% for extracting candidate decision scenes that may contain expert know-how, which is 28 percentage points higher than that of the previous method (e). It can also be seen that the proposed method extracts candidate decision scenes containing expert know-how more accurately than the other baseline methods, namely the previous method + instruction prompt methods (f)–(j), the naive method (l), and the video comparison method (k). In terms of computation time, the proposed method (m) required several minutes, which is of the same order as that of the previous method (e).

Figure 10 shows an example comparison between the similarity map of the naive method (l) and that of the proposed method (m). Ideally, all regions inside the red box are expected to become black, that is, to exhibit low similarity. However, compared with the proposed method (m), the naive method (l) contains scattered white regions, that is, regions of high similarity. In the naive method (l), images from the two videos are directly treated as pairs for judgment. However, the image description generator may generate different descriptions depending on subtle differences between temporally adjacent images, which is considered to cause such scattered high-similarity regions and consequently reduce the extraction accuracy.

Figure 11 shows an example failure case of the proposed method (m) for extracting decision-scene candidates. In this example, the manual-based work video $V_M$ shows the distribution panel door open, whereas the expert work video $V_E$ shows the distribution panel door closed, and therefore this case is expected to be extracted as a candidate decision scene. One reason for this failure is that the prompt input to the VLM during decision-scene candidate extraction includes a prompt for determining whether the two images were recorded at the same location (described in Appendix A-C). Since $V_M$ and $V_E$ were recorded at the same location, this judgment is appropriate with respect to the input prompt. On the other hand, the prompt is not designed to determine differences in the state of the work target, which should be extracted as candidate decision scenes. Therefore, prompt adjustment and improvement methods tailored to the target work situation are considered necessary.

TABLE II
EVALUATION RESULTS ON THE COLLECTED DATA USING THE PROPOSED METHOD AND THE BASELINE METHODS

| Target type | Method ID | Method | VLM | Params. of VLM | $Acc_{union}$ | Inference time |
|---|---|---|---|---|---|---|
| Action | (a) | Previous method [12] | BLIP-2 | 2.7B | 59.0 % | 2 min |
| | (b) | Baseline (video) | Qwen3-VL-Instruct | 8B | 27.0 % | 0.5 min |
| | (c) | Baseline (Naive) | Qwen3-VL-Instruct | 8B | 28.0 % | 36 min |
| | (d) | Proposed method | InternVL2 | 4B | 65.0 % | 6 min |
| Scene | (e) | Previous method [12] | BLIP-2 | 2.7B | 33.0 % | 2 min |
| | (f) | Baseline (Previous+prompt) | BLIP-2 | 2.7B | 28.0 % | 2 min |
| | (g) | Baseline (Previous+prompt) | InstructBLIP | 7B | 28.0 % | 8 min |
| | (h) | Baseline (Previous+prompt) | LLaVA | 7B | 36.0 % | 11 min |
| | (i) | Baseline (Previous+prompt) | InternVL2 | 4B | 31.0 % | 32 min |
| | (j) | Baseline (Previous+prompt) | Qwen3-VL-Instruct | 8B | 37.0 % | 26 min |
| | (k) | Baseline (Video) | Qwen3-VL-Instruct | 8B | 43.0 % | 1 min |
| | (l) | Baseline (Naive) | Qwen3-VL-Instruct | 8B | 46.0 % | 36 min |
| | (m) | Proposed method | Qwen3-VL-Instruct | 8B | 61.0 % | 3 min |

: GT Anomaly frame range
: Predicted Anomaly frame range

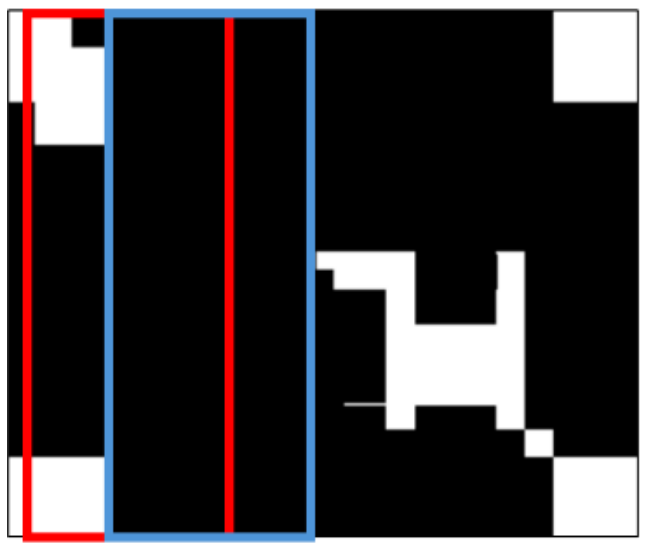

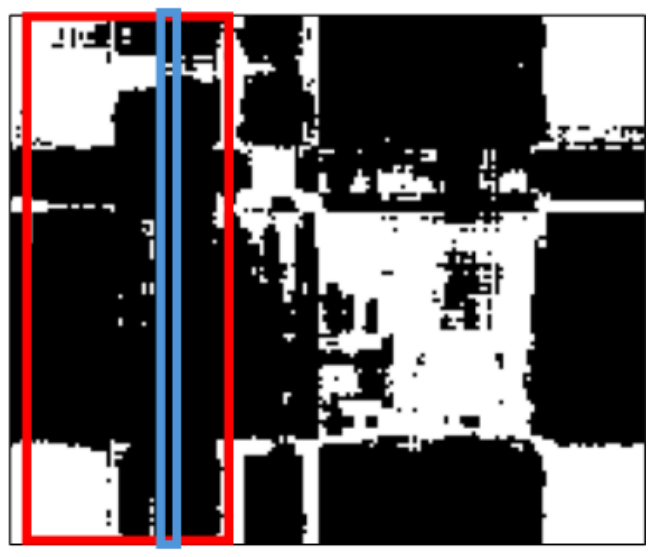

Segment similarity map $S_{smap}$ proposed method (m)

Frame similarity map $S_{fmap}$ Baseline(naive) (l)

**FIGURE 10. Comparison between the segment similarity map $S_{smap}$ of the proposed method (m) and the frame similarity map $S_{fmap}$ of the baseline naive method (l).**

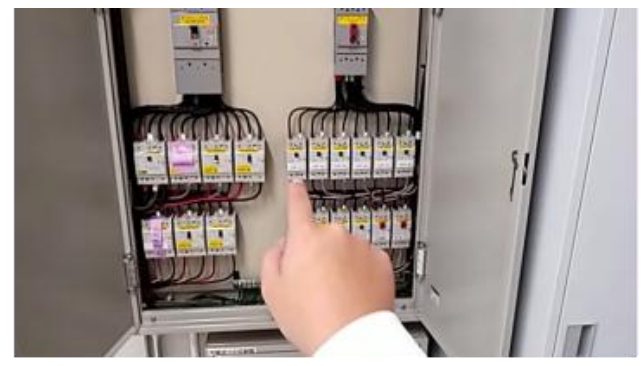

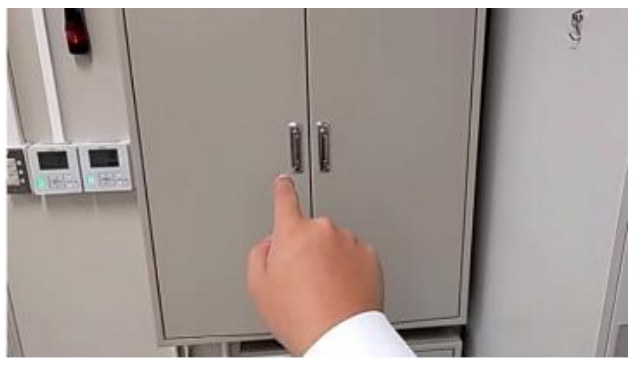

Manual-based work video: $V_M$

Expert's work video : $V_E$

**FIGURE 11. Example of a failure case of the proposed method.**

## VII. CONCLUSION

In the inspection and maintenance of infrastructure facilities, there is a shortage of skilled maintenance workers, and there is a strong need to formalize and transfer the work know-how of skilled workers in order to accelerate the training of new maintenance personnel. In particular, when the target is know-how of which even skilled workers themselves are not consciously aware, it is difficult to extract such knowledge through interviews and similar methods, and therefore a technique is required to efficiently extract candidate know-how from work videos. Conventionally, differences between a manual video and an expert video have made it possible to extract candidate actions corresponding to "what to do," whereas stable extraction of situational judgment know-how corresponding to "where" and "which one to operate" (i.e., decision scenes) has remained a challenge.

In this study, we proposed a method for manual videos and expert videos that extracts candidate actions through inter-video comparison based on image descriptions and extracts candidate decision scenes through comparison at the level of segmented intervals based on intra-video self-similarity. The proposed method realizes the extraction of both candidate actions and candidate decision scenes within a unified framework. Its key feature is that, by extending the comparison unit in decision-scene extraction from the frame level to the interval level, it suppresses both the degradation in extraction accuracy caused by instability in generated descriptions and the increase in computational cost. Based on the above method, we developed a technique for automatically extracting candidate actions and candidate decision scenes that may contain know-how and obtained the following conclusions.

i. Through evaluation using a total of 27 data simulating distribution panel maintenance tasks, we confirmed that the proposed technique operates with computation time comparable to that of the previous method [12].

ii. The extraction rate was 65.0% for candidate actions (+6 points over the previous method) and 61.0% for candidate decision scenes (+28 points over the previous method), both outperforming the previous method.

iii. These results demonstrate that the proposed technique can improve the extraction rates of candidate actions and candidate decision scenes while maintaining computation time, suggesting that it can support the extraction of candidates for formalizing work know-how that even skilled maintenance workers themselves may not be consciously aware of.

Future work will focus on enhancing the accuracy of the proposed method by analyzing actions and decision-making scenes at a finer level of granularity. Furthermore, we plan to develop a technique to detect the reordering of work procedures, as such adjustments often represent a key component of expert expertise. Finally, while the current evaluation was limited to simulated distribution board maintenance experiments, we aim to verify the generalizability of our approach by applying it to a broader range of practical tasks.

## APPENDIX A
## Overview of the Video Comparison Method

An overview of the video comparison method (k), which is one of the comparison methods used in this study, is given below. The video comparison method consists of two steps. In the first step, event detection and event descriptions are output as captions for each of the videos $V_M$ and $V_E$. This method is based on the implementation provided in the official GitHub repository of Qwen3-VL. In the second step, the captions obtained for each video are input to a VLM, which outputs, as captions, which events in one video do not exist in the other video. Based on this output, candidate actions and candidate decision scenes are extracted.

### A. Event Detection

First, using a VLM capable of video captioning, denoted by $M_{\text{vid}}$, event detection and descriptions for the video are generated according to Eqs. (9) and (10):

$$E = M_{\text{vid}}(V, P_{videve}) \tag{9}$$

$$E = \{(start, end, description)_i \mid 0 \leq i \leq K\} \tag{10}$$

Here, $E$ is a set whose elements consist of the detected start time $start$, end time $end$, and description $description$ for each event, and $P_{videve}$ is the captioning instruction prompt. Caption generation is performed for both $V_M$ and $V_E$, producing the captions $E_M$ for $V_M$ and $E_E$ for $V_E$. The same captioning model is used for both $V_M$ and $V_E$.

### B. Anomaly Detection Using Event Detection

Next, extraction of candidate actions and candidate decision scenes using the captions $E_M$ and $E_E$ of $V_M$ and $V_E$ is performed according to Eq. (11):

$$Anomaly\ Event = M_{vid}(E_M, E_E, P_{vidano}) \tag{11}$$

Here, $P_{\text{vidano}}$ is the anomaly-event detection instruction prompt. Anomaly Event is a caption containing information about which event in one video does not exist in the other video. However, because $Anomaly\ Event$ is produced in a free-description format, the output may not always be fixed, making it difficult to process automatically in a unified manner. Therefore, in the last step, a human checks Anomaly Event and performs post-processing required for computing the evaluation metric. For example, when the detected event start and end times are output in seconds, they are converted into frame units.

### C. Prompts Used in This Study

The following prompt was used as $P_{cap}$ in the proposed method (m) described in Section IV-B.

*"You are describing an egocentric view images. Focus on operator's action and the object being acted upon. Describe the concrete action shortly. Avoid mentioning the location or goal - only describe the specific action and target visible in the frame."*

The following prompt was used as $P_{exp}$ in the proposed method (m) described in Section IV-E.

*"Do the two images show the same location? Answer 'Yes' or 'No'."*

The following prompt was used as the instruction prompt for extracting candidate decision scenes in the baseline methods (previous method + instruction prompt), (f) to (j).

*"Describe where you are in detail."*

The following prompt was used as the instruction prompt for extracting candidate actions in the naive method (c).

*"Do the two images show the same Action? Answer 'Yes' or 'No'."*

The following prompt was used as the instruction prompt for extracting candidate decision scenes in the naive method (l).

*"Do the two images show the same location? Answer 'Yes' or 'No'."*

For $P_{videve}$ in Appendix A-A of the video comparison method (b), the following prompt was used for extracting candidate actions.

*"Localize a series of activity events in the video, output the start and end timestamp for each event, and describe each event with sentences. Provide the result in json format with 'mm:ss.ff' format for time depiction."*

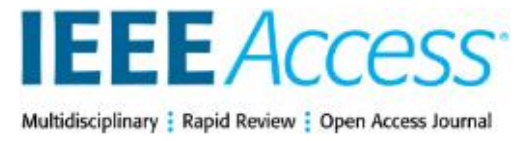

For $P_{videve}$ in Appendix A-A of the video comparison method (k), the following prompt was used for extracting candidate decision scenes.

*"Localize a series of locations in the video, output the start and end timestamp for each locations, and describe each locations with sentences. Provide the result in json format with 'mm:ss.ff' format for time depiction."*

For $P_{vidano}$ in Appendix A-B of the video comparison method (b) and (k), the following prompt was used for both candidate-action extraction and candidate-decision-scene extraction. In practice, $E_M$ and $E_E$, which are the outputs of the VLM in Appendix A-A, are passed to {captions of the reference video} and {captions of the target video}.

*"A series of locations in the reference video:*
*{captions of the reference video}*
*A series of locations in the target video:*
*{captions of the target video}*
*Based on the above two sets of locations, detect anomalies between the reference and target videos. Define anomalies as follows:*

- *"additional anomaly": a locations that appears in the target video but does not exist in the reference video.*
- *"unseen anomaly": a locations that appears in the reference video but does not appear in the target video.*

*Determine which type of anomaly is present, and output **only one type**—either additional anomalies or unseen anomalies, whichever is detected.*
*Do not output both types simultaneously.*
*For the detected anomaly type, output the result in the same JSON format as before, including:*

- *the start and end timestamps ('mm:ss.ff'),*
- ***the anomaly type** ("additional anomaly" or "unseen anomaly"),*
- *a description of the locations,*
- *and a brief explanation of why it is classified as that anomaly type."*

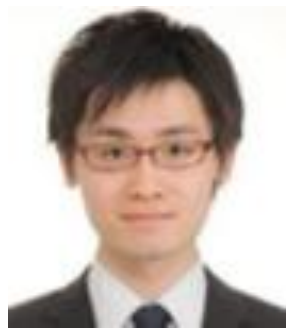

**RYO SAKAI** received Master of Information Science from Osaka University in 2016. He is currently a senior researcher at the Robotics Research Department in Research and Development Group, Hitachi, Ltd. His research interests includes computer vision, machine learning, tacit knowledge extraction, robotics, and intelligent system. He is a member of the Robotics Society of Japan.

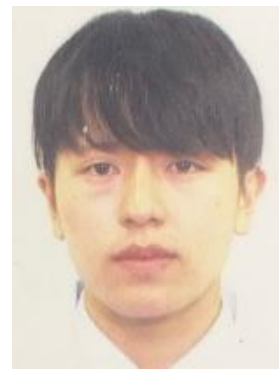

**Kaname Yokoyama** received the bachelor’s degree from Toyota Technological Institute in 2025. He is a master’s student at the Graduate School of Engineering, Toyota Technological Institute, where he is a member of the Intelligent Information Media Laboratory. His research interests lie in computer vision and machine learning.